\documentclass{article}
\usepackage{arxiv}
\usepackage[utf8]{inputenc} 
\usepackage[T1]{fontenc}    
\usepackage{booktabs}       
\usepackage{amsfonts}       
\usepackage{nicefrac}       
\usepackage{microtype}      
\usepackage{lipsum}		
\usepackage{graphicx}
\usepackage{natbib}
\usepackage{doi}
\usepackage{subcaption}
\usepackage{hyperref}
\hypersetup{
    colorlinks=true, 
    linkcolor=blue,  
    filecolor=blue,  
    urlcolor=blue,   
    citecolor=blue   
}

\title{Deep Memory Search: A Metaheuristic Approach for Optimizing Heuristic Search}

\author{\href{https://orcid.org/0000-0002-9936-5987}{\includegraphics[scale=0.06]{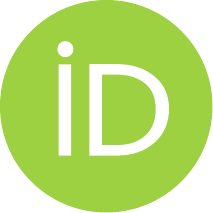}\hspace{1mm}Abdel-Rahman Hedar}\\
Department of Computer Science\\
Faculty of Computers \& Information\\
Assiut University\\
Assiut 71526, Egypt\\
{\tt\small hedar@aun.edu.eg}
\And
\href{https://orcid.org/0000-0002-0003-6914}{\includegraphics[scale=0.06]{orcid.pdf}\hspace{1mm}Alaa E.~Abdel-Hakim} \\
Computer Science Department in Jamoum,\\ Umm Al-Qura University\\
Electrical Engineering Department\\
Assiut University\\
Assiut, Egypt, 71516\\
{\tt\small adali@uqu.edu.sa;}\\
{\tt\small alaa.aly@eng.au.edu.eg}
\And
\href{https://orcid.org/0000-0002-1814-2643}{\includegraphics[scale=0.06]{orcid.pdf}\hspace{1mm}Wael Deabes}\\
Department of Computational, Engineering, Mathematical Sciences (CEMS)\\
Texas A\&M University-San Antonio\\
One University Way, HALL 334\\
San Antonio, TX  78224\\
{\tt\small wdeabes@tamusa.edu}
\And
\href{https://orcid.org/0000-0002-0840-1867}{\includegraphics[scale=0.06]{orcid.pdf}\hspace{1mm}Youseef Alotaibi}\\
Department of Computer Science\\ College of Computer and Information Systems\\ Umm Al-Qura University\\ Makkah 24381, Saudi Arabia\\
{\tt\small yaotaibi@uqu.edu.sa}
\And
\href{https://orcid.org/0009-0002-4879-5298}{\includegraphics[scale=0.06]{orcid.pdf}\hspace{1mm}Kheir Eddine Bouazza}\\
Computer and Information Science Division\\
Higher Colleges of Technology\\
Abu Dhabi, United Arab Emirates\\
{\tt\small kbouazza@hct.ac.ae}
}

\begin{document}
\maketitle

\begin{abstract}
Metaheuristic search methods have proven to be essential tools for tackling complex optimization challenges, but their full potential is often constrained by conventional algorithmic frameworks. In this paper, we introduce a novel approach called Deep Heuristic Search (DHS), which models metaheuristic search as a memory-driven process. DHS employs multiple search layers and memory-based exploration-exploitation mechanisms to navigate large, dynamic search spaces. By utilizing model-free memory representations, DHS enhances the ability to traverse temporal trajectories without relying on probabilistic transition models. The proposed method demonstrates significant improvements in search efficiency and performance across a range of heuristic optimization problems.

\end{abstract}


\section{Introduction}
\label{Sec:Introduction}



Heuristics search is a method to find better optimal solutions at a reasonable computational cost without guaranteeing optimality or feasibility. Metaheuristics algorithms are the set of intelligent strategies used to improve the heuristic procedures efficiency. Metaheuristics algorithms can be considered as an iterative master process used to guide and modify the subordinate heuristics operations to produce high-quality solutions efficiently~\cite{rashedi2018comprehensive, huang2019survey}. Almost all of these algorithms have stochastic behavior and mimics physical or biological processes. They can be classified to
(1) nature-inspired against non-nature inspired,
(2) dynamic against static objective function,
(3) population-based against single-point search,
(4) memory usage against memoryless methods, and (5) single neighborhood against various neighborhood~\cite{beheshti2013review}.

Most of traditional metaheuistic search methods often fail to find optimal solutions due to falling into local optimum.



Machine Learning (ML) techniques enable the computer to learn by inferring some raw data information or knowledge. In Deep Learning (DL), the word “Deep” comes from the contradictory with those commonly used “shallow learning” algorithms such as Support Vector Machine (SVM), boosting, and maximum entropy methods. Typically, the shallow learning algorithms abstract feature by artificial or experiential sampling; thus, the model or network will learn a non-layer structure feature. On the other hand, DL has a deep structure and automatic feature extraction, which retains raw data layer by layer at different levels of abstraction without a field expert human intervention~\cite{LeCun2015Deep, wani2020advances}. Therefore, all these DL advantages are appreciated in representing the nonlinearities, often associated with complex real-world data~\cite{wang2020recent}. 

Introducing DL knowledge in metaheuristic operators such as mutation or crossover operators in Population Management (PM) improves the search process of the metaheuristic methods~\cite{Calvet2017Learnheuristics}. For instance, a coevolutionary GA, proposed in~\cite{handa2002novel}, integrates an extraction mechanism with the crossover operator. The learning concept is presented to the evolutionary computation processes in~\cite{michalski2000learnable}. The Learnable Evolution Model (LEM) uses symbolic learning methods to produce rules that explain why specific individuals are superior to others. Moreover, some ML techniques have been implemented as local searches. For example, authors in~\cite{adra2005hybrid} combined a multi-objective EA with a local search inverse neural network to discover better individuals from previous generations.

Hyperheuristics methods are search algorithms or learning methods for choosing or producing heuristics that solve computational search problems~\cite{ burke2010classification}. Typically, these methods automate the design of heuristic methods and/or deal with a wide range of problems, rather than obtaining better results than problem-specific metaheuristics. Authors in~\cite{burke2013hyper} have conducted a comprehensive survey on hyperheuristics. The potential of associative classifiers in the hyperheuristic approach for solving the training scheduling problem is explored in~\cite{thabtah2008mining}. The classifiers have to indicate the low-level heuristic to use at each step while creating a solution. Reinforcement Learning (RL) and regression techniques are prevalent procedures for selecting heuristics employing an online learning strategy. In~\cite{li2011integrating}, neural networks and logistic regression are employed to predict objective function values of solutions in a hyperheuristic search. 

A hybrid strategy is produced by merging several algorithms/agents which cooperate in parallel or sequentially~\cite{sadeghi2019new, jaafari2019hybrid}. Communication among these algorithms is carried out using either many-to-many (direct) or memory-based (indirect). Regularly, there is an agent that organizes the search of the others, synchronizes the communication. This approach tries to develop a robust methodology that offers high-quality solutions by exploiting each algorithm's specific advantages. For instance, a centralized hybrid metaheuristic cooperative strategy is developed in~\cite{cadenas2009using}, where knowledge is incorporated into the coordinator agent through fuzzy rules. The definition of these fuzzy rules has been acquired based on a knowledge extraction process applied to the results obtained by each metaheuristic. The coordinator agent then collects information and sends orders to each solver agent that will affect its search behavior.

On the other hand, metaheuristics such as GA, PSO, and others have effectively integrated with DL in the past years to optimizing DL tools such as a Convolutional Neural Network (CNN)~\cite{Darwish2020survey}. CNN is a distinct neural network with different weight computation regulations, and it mainly applies iterative formula on error similar to traditional NN for its training processes. Based on the fact that NN (at least part of it) can be coded as a solution vector, the metaheuristics techniques can be applied to optimize their layouts, configurations, weights computations, etc. Integrating metaheuristic in DL speeds up training without declining performance. Metaheuristic has a history of optimizing machine learning models successfully~\cite{Fong2018how, Tian2016survey}.


\section{Modeling Metaheuristics as a Memory Search }

Metaheuristic search can be conceptualized as a memory-based optimization process. By treating the search procedure from a memory-centric perspective, it is possible to dramatically enhance the effectiveness of the search strategy, especially when applied to large and complex search spaces. These spaces are not only vast in spatial dimensions but also in temporal depth. Traditional MDPs typically model temporal trajectories using probabilistic transition systems; however, such models often fail to capture the dynamics of systems that are not governed by explicit state transitions.

This necessitates the development of "model-free" memory representations that can traverse temporal spaces without relying on predefined transition models. One prominent example of such an approach is Reinforcement Learning (RL), specifically in the form of Q-Networks (QNet), which successfully utilize memory-based reward mechanisms to guide learning in environments with uncertain outcomes. RL can thus be viewed as a memory-centric process, where the agent navigates state-action space, alternating between two strategies: exploration and exploitation.

Exploration refers to the investigation of new actions and the subsequent updating of the memory based on the rewards observed. The efficacy of this process relies on non-causal estimates of the reward potential, often represented in deep neural networks, as seen in Deep Q-Learning (DQL). Exploitation, by contrast, prioritizes historical memory data, favoring well-established action-reward mappings.

In this work, we build on this perspective, proposing a novel framework for heuristic search based on a deep memory search mechanism. This method, which we refer to as Deep Heuristic Search (DHS), reimagines heuristic optimization as a layered memory search problem. The following sections detail the methodology behind DHS and its application to various search problems.

In this work, we follow the strategy of modeling metaheuristics as a memory search problem. Therefore, we propose a deep memory search framework to solve the problem of heuristic search and denote it as Deep Heuristic Search (DHS). In the following section, we explain DHS and how it is applied to search problems.







 





\section{Framework: Deep Heuristic Search}
\label{Sec:Framework}
\begin{figure}
    \centering
    \includegraphics[width=0.48\textwidth]{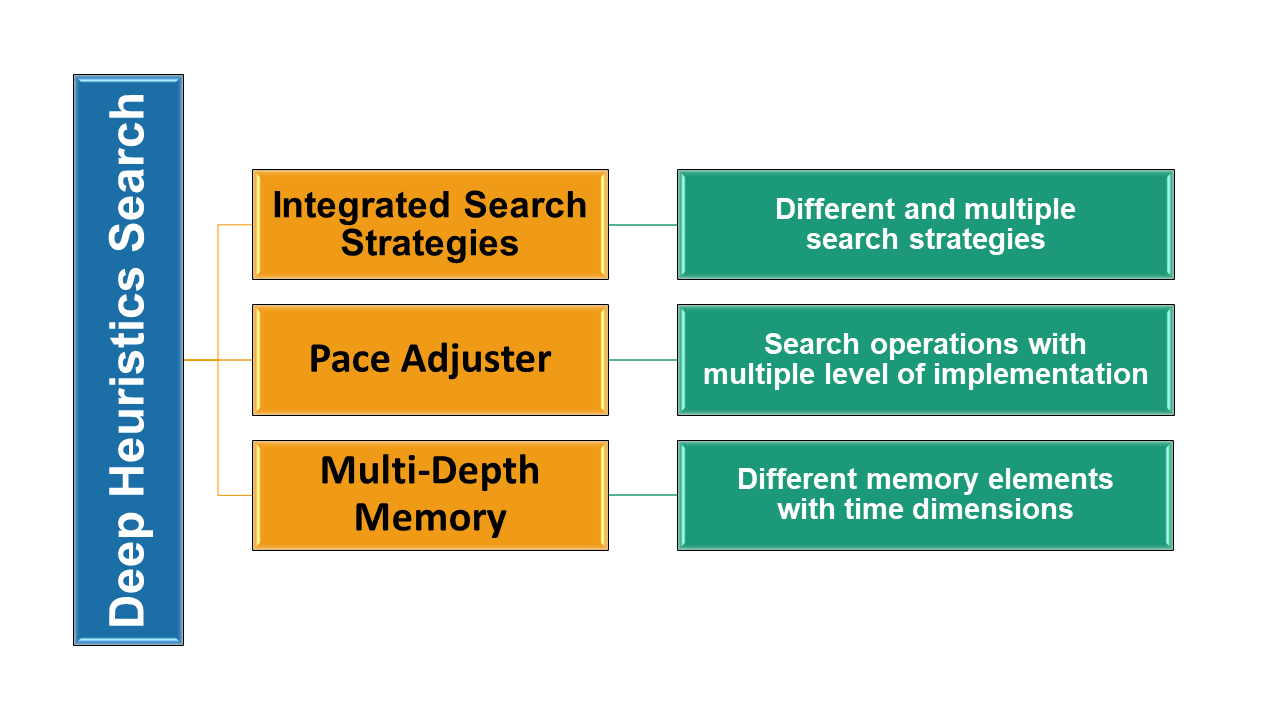}
    \caption{The main components of DHS.}
    \label{fig:DHSfig}
\end{figure}

The Deep Heuristic Search (DHS) is a generic search approach that can extend metaheuristics to perform advanced search. DHS comprises three main components, which are illustrated in Fig.~\ref{fig:DHSfig}. These components can be summarized as follows:
\begin{itemize}
    \item 
    \textbf{Integrated Search Strategies}: invokes different and multiple search strategies including intensification, diversification and hybridization, as well as search restarting strategies.
    \item 
    \textbf{Variant Depth Operations}: define search operations to have multiple levels of implementation according to the search strategies. Each search operation has three implementation levels; expand-mode, normal-mode, and condense-mode. The first and last modes works on expanding or refining the search according to the search strategy used, whether it is diversification or  intensification, respectively. Similar evolution strategies to this proposed pace-adjustable coarse-to-fine paradigm had succeeded in machine learning approaches, e.g. gradient descent for supervised learning~\cite{bottou1998online} and adaptive clustering for unsupervised learning~\cite{hedar2018k,hedar2018modulated}.
    \item \textbf{Multi-Depth Memory}: constructs different memory elements that keep track of historical featured search data along the time dimension. This temporal depth enables the search process to implement the search operations with guidance of recent or historical featured search data according to the search strategies.
\end{itemize}
Deep memory is the core element of DHS. The other components are highly dependent on the operation of the deep memory. Therefore, the operation of deep memory is explained before the other components, as shown in the following subsection.


\subsection{Multi-Depth Memory}

The memory on DHS spans over a two-dimensional space. Traversing along the first dimension provides different degrees of diversity to DHS. This dimension expresses various search features including elitism, visit frequency, characteristics, spatiality and/or recentness. 

\begin{figure}
    \centering
    \includegraphics[width=0.48\textwidth]{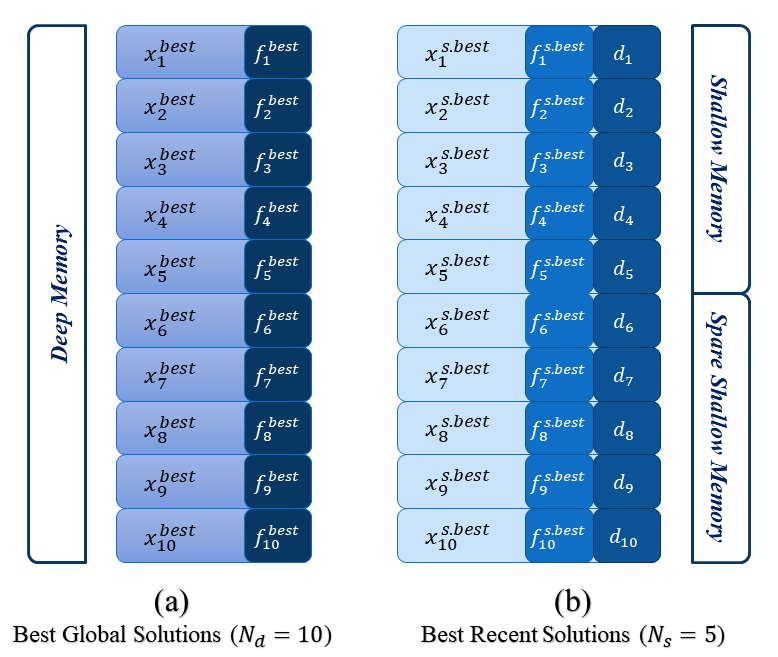}
    \caption{Examples of different memory depth structures in the DHS framework using an elite memory. The size of deep and shallow memories are $N_d = 10$ and $N_s = 5$, respectively.}
    \label{fig:DeepMemory}
\end{figure}

The second dimension gives the DHS memory its deepness property. Mainly, this dimension represents the temporal search history whose controllable depth that is application-dependent. Memory temporal search gives the flexibility to control the application of both local search and global search strategies. Particularly, the DHS memory comprises two levels in terms of depth. The first is a deep one, which is concerned with aged search results. Therefore, this deep memory turns out to serve a global search for featured solutions over the entire search history. The other memory level is a shallow structure, which remembers only recently-visited featured solutions or those which were visited during the current search stage. Fig.~\ref{fig:DeepMemory} shows how the deep and the shallow structures are constructed in the elite DHS memory. This memory elements collects predefined numbers of best obtained solutions as explained later. In the deep memory structure, the $N_d$ best solutions which are globally-reached by the overall search are stored with their objective function values. The shallow memory structure preserves the $N_s$ best recently-visited solutions. Practically, $N_d$ and $N_s$ are selected such that $N_d > N_s$. 

The depth of memory elements consists of two dimensions, one dimension represents diversity and the other does for the time. Therefore, in order to maintain the diversity, DHS memorizes different search features including elitism, visit frequency, characteristics, spatiality and/or recentness. Moreover, in order to control the application of both local search and global search strategies, the DHS creates a time depth in memory. In other words, each memory element can have two structures, one deep and another shallow one. The deep structure memorizes the search featured solutions globally over the entire search history, while the shallow structure only remembers recently visited featured solutions or those which were visited in the current search stage. For example, in Fig.~\ref{fig:DeepMemory}, assume $N_d = 10$ and $N_s = 5$. In this case, the construction of deep and shallow structures in the elite memory are illustrated. In the deep memory structure, the ten best solutions obtained overall the search are stored with their function values. On the other side, the shallow memory structure, the only five best recently-visited solution are stored in this memory. To control the temporal depth in the shallow memory structure, the depth index ($d_i, i = 1, \ldots,5$) is used to track the number of iterations or generations that have passed since visiting the corresponding solutions. To prevent any sudden drop in the number of elements in shallow memory, the extended shallow memory is used to preserve the best next elements in the shallow memory.

For the diversity dimension, we adopt several diversity approaches.  In the next subsections, we discuss the most important types of these diversity methodologies. 

\subsubsection{Elitism}
In this type of memory, the elite solutions are stored based on their objective function values. Therefore, the best reached solutions at a moment are stored in the deep elite memory. The best-visited solutions in the last ($d_e$) iterations/generations or in the current search stage are stored in the shallow elite memory, as well as its extended memory.

\subsubsection{Visit Frequency}
The accumulated number of solutions or region visits is stored in the deep frequent memory. 
On the other side, the shallow frequent memory and its extended peer store only the number of visits in the last ($d_f$) iterations/generations or in the current search stage.

\subsubsection{Characteristics}
Some characteristics of featured solutions can be saved to guide the search process. For example, connected sub-graphs can help in search for connected dominating sets in network applications. Therefore, storing the best connected solutions can direct the research process towards finding better solutions. Therefore, the deep characteristic memory stores the best featured solutions obtained so far based on selected characteristics. The shallow characteristic memory and its extended memory only store the best featured solutions obtained in the last ($d_c$) iterations/generations or in the current search stage.

\subsubsection{Spatiality}
The search space can be sampled or partitioned and some landmarks inside it or in its partitions can be stored. These spatial data of the search space helps introducing new and diverse solutions. Moreover, generating solutions in some applications needs composing solutions that covers different regions of the search space. The deep spatial memory stores the space landmark data and their visiting times. The shallow spatial memory and its extended memory only store the search data for the visited landmarks in the last ($d_s$) iterations/generations or in the current search stage.

\subsubsection{Recentness}
A tabu list is constructed to contain the ($d_r$) last visited solutions. This list keeps the DHS aware of the most recently visited solutions. Therefore, the unnecessary, yet exhausting irrelevant, return to those solutions is avoided.

\subsection{Pace Adjuster}

Within the operation of the proposed DHS, a traditional metaheuristics methodology is invoked and directed by the DHS components. DHS has three different modes of operations: \textit{normal, expand,} and \textit{condense} modes. In the normal mode, the search operations used in the invoked metaheuristic is applied as its original definitions. If some search strategies are called, then the definitions of the operations should be modified to fulfill the considered search strategies. The expand and condense implementation modes are defined to be used during the diversification and intensification, respectively. More details with illustrating examples about these search modes are given in the following. 

\subsubsection{Normal-Mode Operations}
Metaheuristics have their own search operations and they are defined to fulfill the invoked optimal search strategies. In this work, three well-known search operations have been considered to illustrate the new concept of deep operations. However, the deep operation concept can be extended to adopt other search operations in different metaheuristics. The considered three search operations are shown in the following. 

\begin{itemize}
    \item \textbf{The arithmetic crossover in genetic algorithms.} Given two parents; $x_1$ and $x_2$, then their children can be generated as:
    \begin{eqnarray} 
    \label{eq:xover}
        y_1 &=&  \lambda x_1 + (1-\lambda)  x_2, \\ \nonumber
        y_2 &=&  (1-\lambda) x_1 + \lambda  x_2,
    \end{eqnarray}
    for some $\lambda \in [0,1]$.
    \item \textbf{The self-adaptation mutation in evolution strategies.} A mutated child $(y,\theta)$ can be obtained from a parent $(x,\sigma)$, using the following equations, which assigns the $i$-th component of the mutated child as:
  \begin{eqnarray}
  \label{eq:muta}
    \theta_i &=& \sigma_i ~e^{\tau' N(0,1)+\tau N_i(0,1)},  \\ \nonumber
    y_i &=& x_i + \theta_i N_i(0,1), 
  \end{eqnarray}
where $\tau' \propto 1/\sqrt{2n}$ and $\tau \propto 1/\sqrt{2\sqrt{n}}$. These coefficients are usually set to one.
    \item \textbf{The neighborhood zones in tabu search.} Neighborhood trial moves $\{x_1,x_2,\ldots,x_m\}$ of the current iterate solution $x$ can be computed inside the following zones,
    \begin{equation}
    \label{eq:neighbor}
        (i-1) r < \|x_i-x\|   \leq i r,
    \end{equation} 
    for some neighborhood radius $r$, and $i=1,2,\dots,m$.
\end{itemize}
\noindent Implementing the above-mentioned operations as defined is known as normal-mode operations in the DHS framework. 

\begin{figure}
    \centering
    \includegraphics[width=0.48\textwidth]{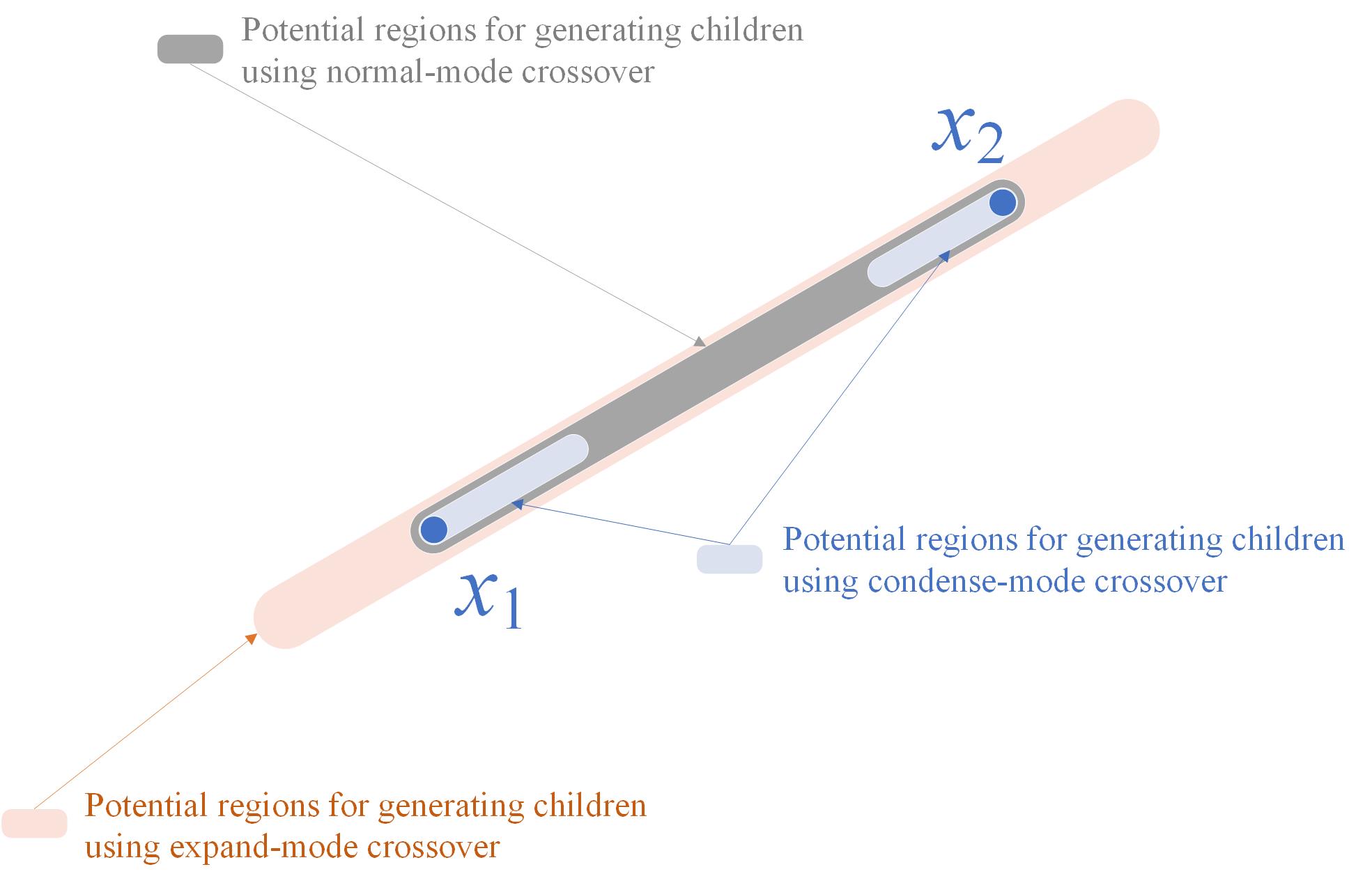}
    \caption{Operation modes for a crossover operator.}
    \label{fig:xoverMode}
\end{figure}
\begin{figure}
    \centering
    \includegraphics[width=0.48\textwidth]{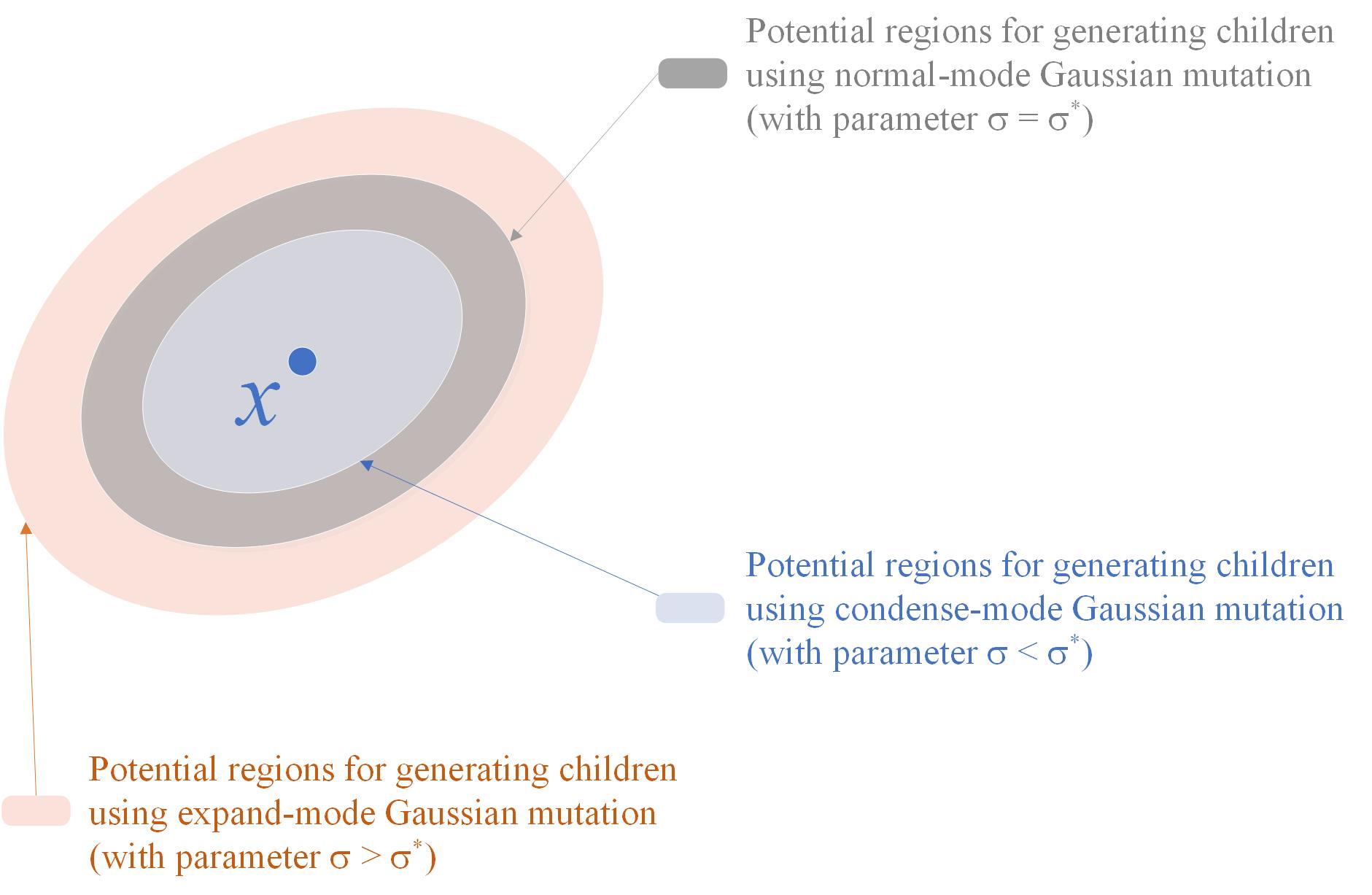}
    \caption{Operation modes for a mutation operator.}
    \label{fig:mutaMode}
\end{figure}
\begin{figure}[t]
    \centering
    \includegraphics[width=0.9\textwidth]{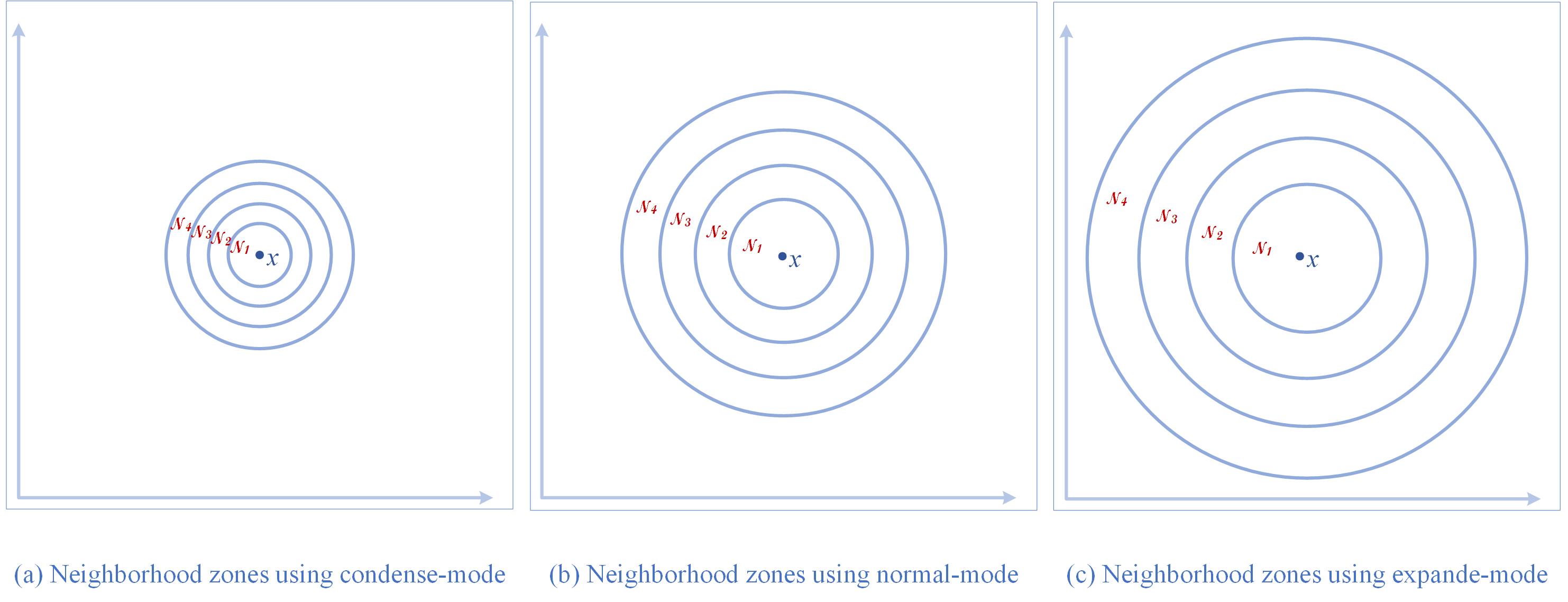}
    \caption{Operation modes for constructing neighborhood zones.}
    \label{fig:neighborMode}
\end{figure}

Examples of such mode are shown in gray in Fig.~\ref{fig:xoverMode}, Fig.~\ref{fig:mutaMode}, and in Fig.~\ref{fig:neighborMode}:b.

\subsubsection{Expand-Mode Operations}
In the DHS framework, Expand-Mode operations are defined to stretch the search regions of interest, and hence allow generating more diverse solutions in order to fulfil a wider diversification process. For instance, this mode can be applied for the aforementioned considered operations as follows:
\begin{itemize}
    \item \textbf{Arithmetic crossover:} set the parameter $\sigma > \sigma^*$, where $\sigma^*$ is the value of the parameter if the normal-mode is implemented.
    \item \textbf{Self-Adaptation Mutation:} set the parameter $\lambda \in [-1,1]$ instead of setting it in $[0,1]$.
    \item \textbf{Neighborhood Zones:} set the neighborhood radius $r > r^*$, where $r^*$ is the value of the radius in the normal-mode.
\end{itemize}
Illustrative examples of such mode are shown in light orange in Fig.~\ref{fig:xoverMode}, Fig.~\ref{fig:mutaMode}, and in Fig.~\ref{fig:neighborMode}:c.

\subsubsection{Condense-Mode Operations}
Condense-Mode operations are used to enable the DHS search process to focus the search process at finer regions in order to fulfil an exhaustive intensification process. The condense-mode of operations can be applied on the aforementioned operations as follows:
\begin{itemize}
    \item \textbf{Arithmetic crossover:} set the parameter $\sigma < \sigma^*$, where $\sigma^*$ is the value of the parameter if the normal-mode is implemented.
    \item \textbf{Self-Adaptation Mutation:} set the parameter $\lambda \in [0.5,1]$ instead of setting it in $[0,1]$.
    \item \textbf{Neighborhood Zones:} set the neighborhood radius $r < r^*$, where $r^*$ is the value of the radius in the normal-mode.
\end{itemize}
Examples of such mode are shown in light blue in Fig.~\ref{fig:xoverMode} and Fig.~\ref{fig:mutaMode}, and in Fig.~\ref{fig:neighborMode}:a.

\subsection{Integrated Strategic Search}
\begin{figure}
    \centering
    \includegraphics[width=0.95\textwidth]{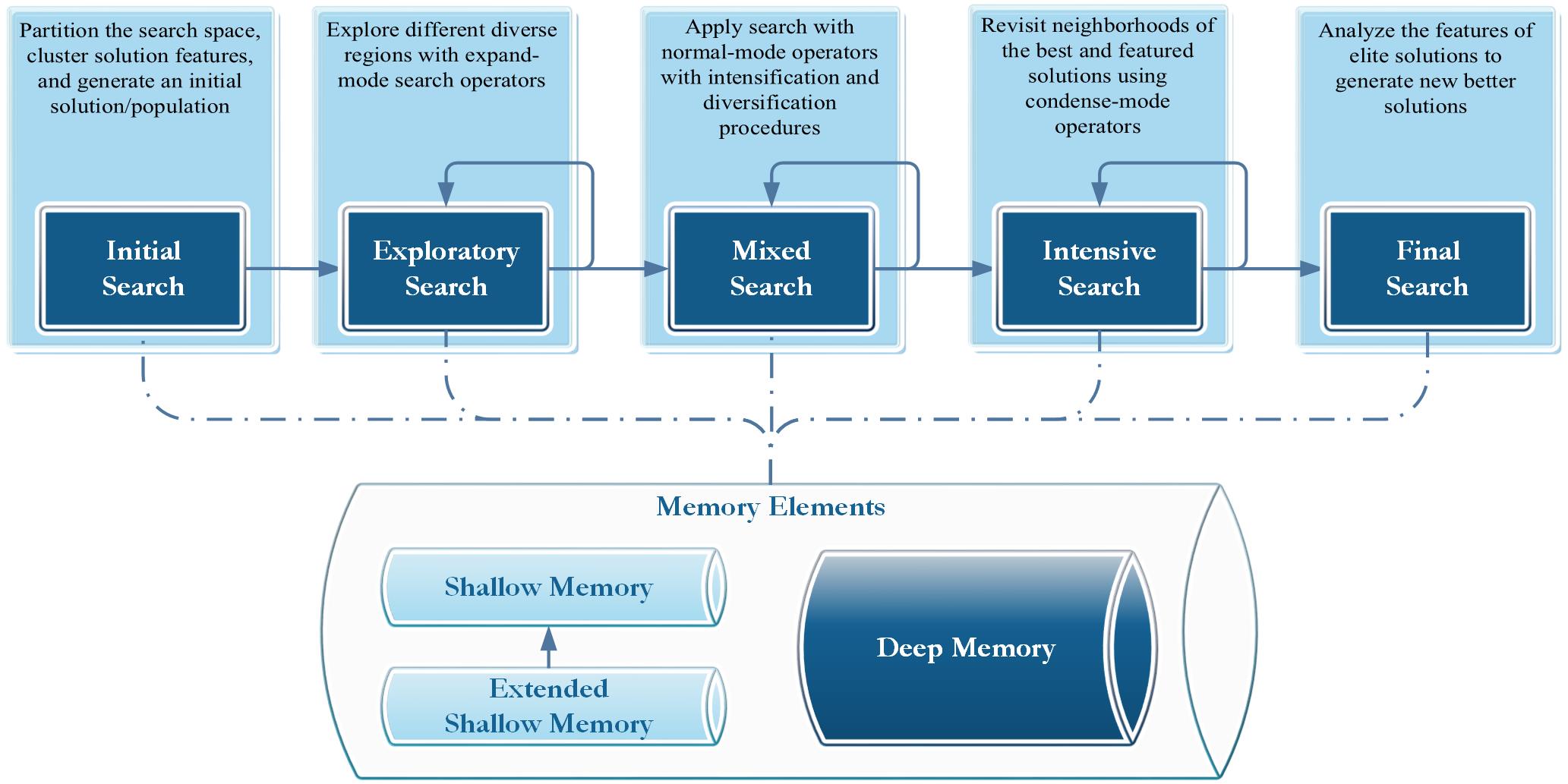}
    \caption{The detailed layout of the DHS framework.} 
     \label{fig:DHS}
\end{figure}

The deep search is the main component of the DHS framework which enables five level of search stages. Each search stage has its own strategies to accomplish the search goals. The main layout of the DHS framework is given in Fig.~\ref{fig:DHS} which contains the five deep search stages, which are:
\begin{itemize}
    \item Initial Search
    \item Exploratory Search
    \item Mixed Search
    \item Intensive Search
    \item Final Search
\end{itemize}
The first and last stages are special ones, which seek initializing and finalizing the search process, respectively. The other search stages are extensive diversification and/or intensification search with restarting strategies. Since the DHS extends one or more of metaheuristic methods, then their search strategies are updated to fill up the five DHS search stages.

\subsubsection{Initial Search} 
The main objective of the Initial Search is to generate an initial solution set and construct the initial memory structures. Therefore, in this search stage special search strategies are invoked to investigate the main features of the search space. Specifically, the Initial Search works on partitioning the search space and clustering solution features as well as initializing the search memory. Then, a large solution set called the \textit{\textbf{gentry}} is generated to cover different and diverse space partitions and solution clusters. Therefore, the deep spatial memory can help in constructing such partitions and clusters. Finally, an initial solution/population can be selected among the best candidates in the gentry. 

\subsubsection{Exploratory Search} 
In the exploratory search stage, the search operations of the considered metaheuristics are modified to generate more diverse solutions. The main strategies of this search stage is how to perform a quick wide exploration process of the search space and to construct more concrete search memories about this space. Therefore, expand-mode operations are defined to fulfil this strategies. Moreover, a restarting mechanism can be implemented to adeptly configure such expand-mode operations. Some search memory elements can be used to examine the exploratory search progress in order to finish it and move to the next search stage.

\subsubsection{Mixed Search}
The search operations of the considered metaheuristics are applied with their normal-mode definitions in the mixed search stage. However, special intensification and diversification procedures can be inlaid whenever promising solutions are detected or the research process fails to find better solutions, respectively. The mixed search can be restarted with new adapted control parameters especially whenever the diversification process suggests new diverse solutions. The deep spatial and frequent memories can define a practical automatic termination criteria to stop this main search stage.

\subsubsection{Intensive Search} 
In the intensive search stage, the best candidates stored in the deep elite and featured memories are revisited using refiner search process. Therefore, neighborhood regions of these best and featured solutions are searched using condense-mode operations. The search process maybe restarted from the same solution or from a new one with adeptive configuration of the search control parameters.

\subsubsection{Final Search} 
This final search stage is a special search that collects the final forms of the best reached solutions. Then, the components of these solutions are analyzed in order to generate new better solutions. Such analysis process depends on the problem structure and objective. For example, the presence of similarities in some components of the best solutions can be used as a core for generating new solutions.















\section{Conclusion}
\label{Sec:Conclusion}

The Deep Heuristic Search (DHS) framework offers a novel memory-based approach to solving metaheuristic optimization problems. By leveraging multiple search layers and memory-guided exploration-exploitation strategies, DHS provides an efficient means of navigating large and complex search spaces. The method's ability to operate without probabilistic transition models makes it particularly well-suited to problems where traditional approaches fall short. DHS represents a promising step forward in the development of deep memory-based heuristic search methodologies across diverse fields.




\bibliographystyle{unsrtnat}  

\end{document}